# LETTING THE CAT OUT OF THE BAG: GENERATION FOR SHAKE-AND-BAKE MT

CHRIS BREW
Sharp Laboratories of Europe Ltd.
Oxford
30 April 1992
Chris.Brew@ed.ac.uk

## 1. Introduction

This paper discusses an implementation of the sentence generation component of a Shake-and-Bake Machine Translation system.. Since the task itself is NP-complete, and therefore almost certainly intractable our algorithm is a heuristic method based on constraint propagation. We present preliminary evidence that this is likely to offer greater efficiency than previous algorithms.

In SLE's approach to multilingual machine translation. [Whitelock, 1991, this conference] we envisage the process of sentence generation as beginning from a *multiset* or *bag* of richly structured *lexical signs* rather than from a conventional logical form or other underlying structure. The translation equivalences are stated between sets of lexical signs, with the superstructure of non-terminal symbols being no more than the means by which monolingual grammars are implemented

The work described here was motivated by a desire to improve on a correct but inefficient algorithm provided by Whitelock [Whitelock, 1991, this conference]. We begin by introducing the problem, proceed by investigating its worst-case behaviour, and conclude by describing new algorithms for Shake-and-Bake generation.

Since the linear order of the source language is not transferred into the bag, it is the business of the monolingual grammar writer to ensure that the word-order requirements of the target language are suitably encoded, and the business of the algorithm designer to ensure that this encoding is exploited as efficiently as possibl.. For an example of the grammar writer's responsibility, the difference between "Mary likes Frances" and "Frances likes Mary" can be encoded in the sharing of index variables between the proper nouns and the verb. For an example of the algorithm designer's responsibility, it would be a mistake (as Whitelock has noted) to provide a translation or generation algorithm which unintentionally unified the two index variables, leading to a reading in which "Mary" and "Frances" are alternative names for the same person.

## 2. Shake-and-Bake Generation

### 2.1 Complexity results

#### 2.1.1 Specification

Shake-and-Bake generation has more in common with a parsing algorithm than with conventional generation from logical form or other underlying structure. The input to the task consists of the following elements:

- A set (**B**) of lexical signs having cardinality |**B**|.
- A grammar (**G**) against which to parse this input string.

and a solution to the problem consists of

- A parse of any sequence (**S**) such that **S** contains all the elements of **B**.

The unordered nature of **B** is the difference between Shake-and-Bake generation and conventional CFG parsing. Although we are really interested in more expressive grammar frameworks, it will for the moment be convenient to assume that **G** is a simple context-free grammar. Since it is always possible to re-implement a CFG in the more expressive formalisms, the Shake-and-Bake generation problem for these formalisms is certainly at least as hard as the equivalent problem for CFGs

### 2.1.2. Upper bound

Simply stating the Shake-and-Bake problem in these terms yields a naive generation algorithm and a minor technical result. The algorithm, which we shall call `generate-and-test,` is simply to feed the results of permuting the input bag to a standard context-free parser. The minor technical result, which will used to establish a complexity result in §2.1.4., is that Shake-and-Bake generation is in NP. Once we note that

• Context-free parsing is a polynomial process.

• The "magical" non-determinism which NP allows is enough to permute the input string using no more than polynomial time and space.

it becomes obvious that Shake-and-Bake generation falls within the definition of NP given by Garey and Johnson [Garey and Johnson, 1979, p 32]. This provides an upper bound on the complexity of Shake-and-Bake generation by showing it to be in NP (rather than being, for example, PSPACE hard or worse). All that remains to be shown is whether it also satisfies the definition of NP-completeness given on p38 of the same work.

### 2.1.3 Lower Bound

The purpose of this section is to establish a lower bound on the complexity of Shake-and-Bake generation. We do this by demonstrating that Shake-and-Bake generation is equivalent to the problem which Garey and Johnson [Garey and Johnson, 1979, pp 50-53] call THREE-DIMENSIONAL MATCHING, but which we prefer to refer to as the MENAGE A TROIS PROBLEM. This is a generalization to three dimensions of the well-known MARRIAGE PROBLEM. In the MARRIAGE PROBLEM the task is a constrained pairwise matching of elements from two disjoint sets, while in the MENAGE A TROIS PROBLEM, the task is the construction of triples based on elements from three disjoint sets. While the original two-dimensional problem is soluble in polynomial time, the three-dimensional analogue is NP-complete. It is therefore of interest to demonstrate a reduction from MENAGE A TROIS to the Shake-and-Bake generation problem, since this serves to establish the complexity class of the latter problem.

### 2.1.4 The MENAGE A TROIS in the bag

The MENAGE A TROIS problem involves three sets **A, B, C** of identical cardinality **n**, having elements which we shall refer to as $a_1...a_n$, $b_1...b_n$ and $c_1...c_n$, along with a set **M** of *constraints* each of which is a triple which represents a mutually acceptable *ménage à trois* . The overall goal is to find a set of three-way marriages selected from M such that every member of A, B and C participate in exactly one triple. Garey and Johnson provide a proof, after Karp, that the MENAGE A TROIS problem is equivalent to the standard problem of 3SAT. We now provide a polynomial-time reduction from an arbitrary instance of MENAGE A TROIS to an instance of Shake-and-Bake generation, which allows the same conclusion to be drawn for this problem.

We start by forming an input string **S** containing all the elements of the three sets **A,B,C**, in any order. We then construct a context-free grammar **G** from **M**, such that each constraint of the form {a<sub>i</sub>,b<sub>j</sub>,c<sub>k</sub>} corresponds to a distinct ternary production in **G**, with the form

`x --> a_i,b_j,c_k`.

To complete the grammar we need a final production of the form

`x --> x, x.`

The role of this production is to ensure that a parse can be achieved if and only if there is a way of covering the input string with constraints. The construction of the grammar and the input string is clearly a polynomial process. Context-free parsing has the property that a leaf node of the input string can only be directly dominated by one node of the final analysis tree, and by the definition of Shake-and-Bake generation given above the Shake-and-Bake process for `G` and `S` must succeed if and only if `G` admits, under the standard node-admissibility interpretation of context-free grammars, a string `S1` which is a permutation of `S`. By combining the preparation described above with Shake-and-Bake generation, we obtain a solution of MENAGE A TROIS. Taken together with the result from §2.1, this constitutes a demonstration that Shake-and-Bake generation is NP-complete.

### 2.2. Conclusion

It is highly unlikely that we will be able to find algorithms for Shake-and-Bake generation which are efficient in the general case: while it might conceivably turn out that NP-complete problems are, after all, soluble in polynomial time, they must for the moment be assumed intractable. We therefore proceed to the discussion of algorithms which are exponential in the worst case, but which do not necessarily exhibit the exponential behaviour unless the grammar is extremely unusual.

## 3. Improved Generation algorithms

It may not be possible to find algorithms which come with a useful theoretical upper bound on run-time cost, but it is still worth looking for ones which will provide acceptable behaviour for realistic inputs. This makes the assessment of such algorithms an empirical matter.

### 3.1. Whitelock's algorithm

Whitelock's algorithm is a generalisation of Shift-Reduce parsing. It is an improvement on the naive **generate-and-test** outlined above, but exhibits exponential behaviour even on the sort of inputs which our MT system is likely to encounter.

A case in point is found in the analysis of English adjectives. We shall be using the phrase "The fierce little brown cat" as our main example.

|  |
|---|
| The fierce brown little cat |
| The brown fierce little cat |
| The brown little fierce cat |
| The little brown fierce cat |

Figure 1

For the sake of argument suppose that we need to rule out the questionable versions of the phrase in Figure 1. It is not clear that these phrases are completely ungrammatical, but they serve the present purpose of providing an illustration, and most English speakers agree that they would only be produced in highly unusual circumstances.

In order to cover this data in a unification grammar, we adopt the encoding shown in Figure 2. This states that "fierce" must precede "little" or "brown" if either of these are present, that "little" must precede "brown" if both are present. (The type assignments are based on the systematic encoding of a finite-state machine.)

| Item | Remainder | Active Part |
|---|---|---|
| the | np / | n(_) |
| fierce | n([]) / | n([1 \| _]) |
| little | n([1]) / | n([1,1 \| _]) |
| brown | n([1,1])/ | n([1,1,1 \| _]) |
| cat | n(_) | <none> |

Figure 2

This set of type assignments prevents the dubious phrases listed in Figure 1, but still allows syntactically acceptable phrases such as "The fierce cat", "The little cat" and "The little brown cat". In principle, this means that generation from a bag produced by analysis of "La petite chatte féroce et brune" will eventually yield the correct outcome. Unfortunately, for phrases like this one Whitelock's algorithm displays spectacular inefficiency.

|  |
|---|
| The fierce brown cat |
| The fierce cat |
| The brown cat |
| The little cat |
| The cat |

Figure 3

For example, the algorithm will construct the intermediate phrases shown in Figure 3, all of which eventually lead to an impasse because it is impossible to incorporate the remaining adjectives while respecting the prescribed ordering. The reason for this behaviour is that Whitelock's algorithm makes reductions on the basis of mere possibility, rather than taking account of the fact that all elements of the input bag must eventually be consumed.

### 3.2. Constraint propagation

We are looking for a global property of the input bag which can be exploited to prune the search space involved in the generation process, and we wish to exploit the completeness property which Whitelock's algorithm neglects. Van Benthem's [1986] observation that categorial grammars display a count invariant, while promising, cannot be directly applied to unification based grammars. As an alternative we develop an approach to Shake-and-Bake generation in which the basic generator is augmented with a simple constraint propagation algorithm [Waltz, 1972]. The augmented generator is potentially more efficient than Whitelock's, since the constraint propagation component helps to direct the generator's search for solutions.

## 4.The algorithm
### 4.1.Description of the algorithm

Our new algorithm relies on the ability to break a bag of signs into its component basic signs, and arranges these signs according to their nesting level. **Nesting level** is defined to be zero for the functor of a categorial sign, one for the functors of its direct arguments, two for the functors of any arguments which form part of these arguments, and so on. Thus the category **a/(//c)/d** has an **a** with nesting level 0, a **b** and a **d** with nesting level 1, and a **c** with nesting level 2. We organize the basic signs of the input bag into a graph in which two nodes are linked if and only if

- Their nesting levels differ by exactly one.
- They arise from different lexical items.

These are necessary but not sufficient conditions for two basic signs to undergo unification in the course of a completed derivation.

```
Node      Category          Lexical Item       Nesting
0         np                : <dummy>           1
1         np                : the               0
2         n(_)              : the               1
3         n([])             : fierce            0
4         n([1|_])          : fierce            1
5         n([1])            : little            0
6         n([1,1|_])        : little            1
7         n([1,1])          : brown             0
8         n([1,1,1|_])      : brown             1
9         n([1,1,1])        : cat               0
```
Figure 4

In the example of the fierce brown cat we obtain connections listed in Figure 4 and the graph shown in Figure 5

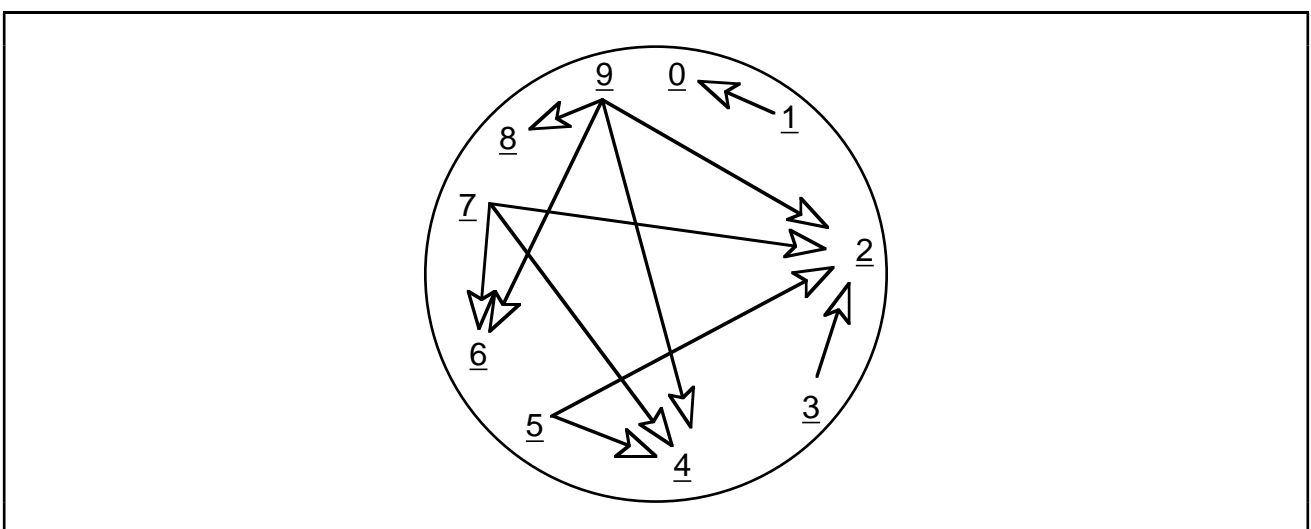
Figure 5

It simplifies the algorithm to hallucinate a dummy node corresponding to the "inverse" of the target category of the derivation ; this is node 0. The node numbers shown in Figure 5ff. correspond to those listed in Figure 4 .The structure is a directed graph, in which elements are linked if and only if they *may* stand in a functor/argument relationship.

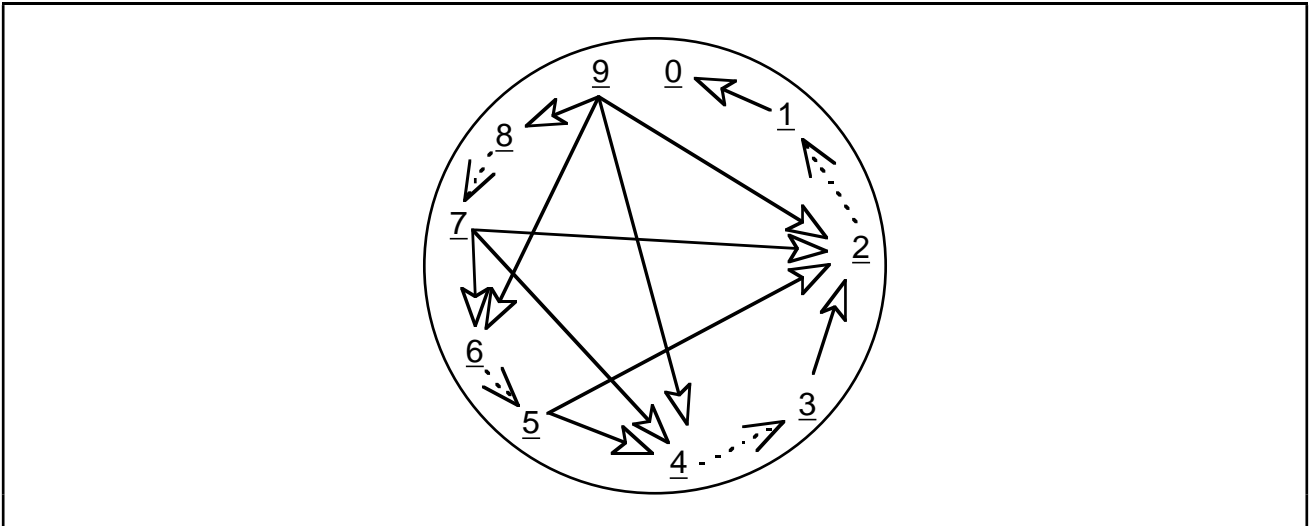

Figure 6

The results of doing this are shown in Figure 6. The task of parsing is reinterpreted as a search for a particular sort of spanning tree for the graph. Our new algorithm is an interleaving of Whitelock's shift reduce parsing algorithm with a constraint propagation component designed to facilitate early detection of situations in which no suitable spanning tree can be built. This helps to prune the search space, reducing the amount of unnecessary work carried out during generation.

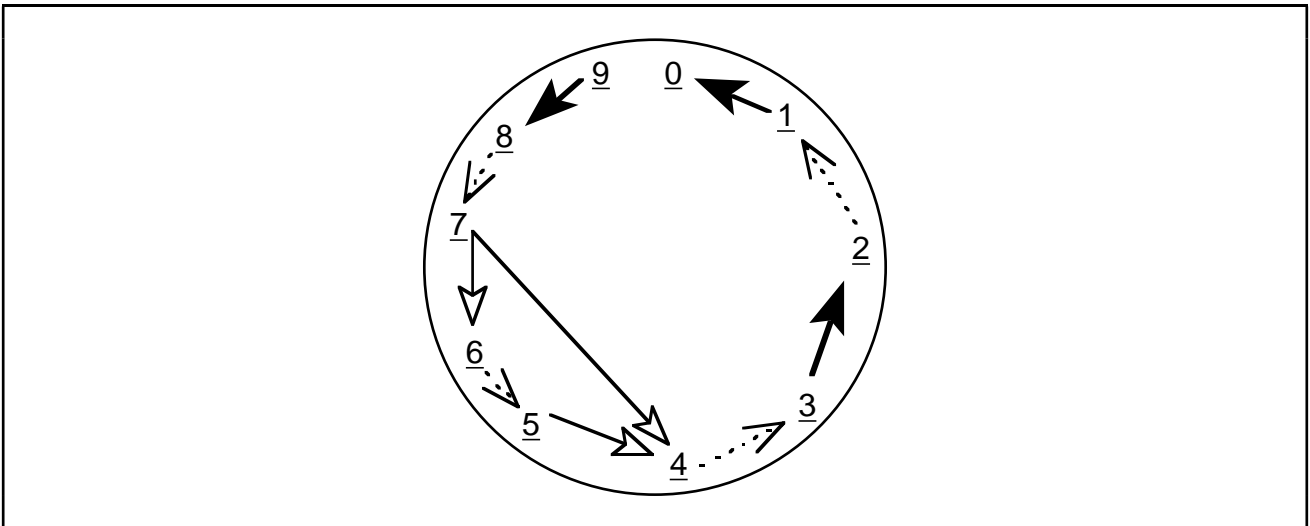

Figure 7

We know that each node of the structure must participate in exactly one functor/argument relationship, but in order to distinguish between those elements which *may* stand in such a relationship and those which actually form part of a complete solution, it is necessary to propagate constraints through the graph. In order to do this it is convenient to add construction lines linking signs in functor position to the corresponding signs which occur in their argument positions.

In Figure 6 we can immediately see that node 3 must be connected to node 2, since there are no other links leading away from node 3. Similarly the link from 9 to 8 must be present in any spanning tree, since there is no other way of reaching node 8. Node 1 must be connected to node 0 for analogous reasons.

Once these links have been established, we can delete alternative links which they preclude. This results in the deletion of the lines from node 9 to nodes 6, 4 and 2, and that of the line from 7 to 2. This produces Figure 7.The resulting system can once again be simplified by deleting the line from node 7 to node 4, yielding a unique circuit through the graph. This corresponds to the correct analysis of the phrase in question. In this example the constraints encoded in the graph are sufficient to drive the analysis to a unique conclusion, without further search, but this will not always happen. We need a combination of constraint propagation with a facility for making (reasonably intelligent) guesses when confronted with a choice of alternatives. This is described in the next section.

### 4.2.The code

We combine the constraint propagation mechanism with Whitelock's original shift-reduce parser, propagating constraints after every reduction step. The parser has the role of systematically choosing between alternative reductions, while the constraint propagation mechanism fills in the consequences of a particular set of choices.

Listing 1 provides a schematic Prolog implementation of the algorithm described in this section. The code is essentially that of a shift-reduce parser , with the following modifications

• One of the elements in a reduction is taken from the top of the stack, while the other is taken from anywhere in the tail of the stack. This idea, due to Whitelock and Reape, ensures that the input is treated as a bag rather than a string.

• At initialization a constraint graph is constructed. Every time a reduction is proposed the constraint propagation component is informed, allowing it to (reversibly) update the graph by propagating constraints. Constraint propagation may fail if the constraint mechanism is able to show that there will be no way of completing a suitable spanning tree given the choices which have been made by the shift-reduce component.

In this algorithm it is the role of the shift-reduce component to make guesses, and the role of the constraint solver to follow through the consequences of these guesses. In the limit this will clearly reduce to an inefficient implementation of exhaustive search, but this should not be a surprise given the NP-completeness of the task.

### 4.3.Results

We have conducted an experiment to show the relative performance of the two algorithms. Figure 8 shows the number of reductions which were carried out by each algorithm in dealing with a range of sentences about fierce cats and tame foxes (The talk of cats and bags is because we are trying to get a CATegory out of a BAG. The constraint propagation algorithm attempts substantially fewer reductions than the original in all cases, with an increasing performance advantage for longer sentences. This remains true even when real-time measurements are used, although the difference is less marked because of the overhead of the constraint propagation algorithm.

```
generate(BagIn,Result) :-
    new_constraint_graph(BagIn,G),
    shake_and_bake([],Result,BagIn,[],G).

 shake_and_bake([Sign],Sign,[], [],_) % termination

shake_and_bake(P0,Sign, [Next|Bag0], Bag,G):- % shift
    push(Next, P0, P)
    shake_and_bake(P,Sign,Bag0,Bag,G).

 shake_and_bake(P0,Sign,Bag0,Ba,,G) :- % reduce
    pop(P0, P1),
    delete(Second,P1,P2),      % treat input "string" as a bag
    unordered_rule(Mom,First,Second,Info),   % 4th arg is info for the
                                             % constraint propagation.
    update(Info,G),                          % constraint propagation
    push(Mom, P2, P),
     shake_and_bake([P,Sign, Bag0, Bag,G).
```

Listing 1

| Example | Length | Whitelock | Constraint Propagation |
|---|---|---|---|
| 1 a fox | 2 | 1 | 1 |
| 2 a yellow fox | 3 | 3 | 2 |
| 3 a tame yellow fox | 4 | 7 | 3 |
| 4 a big tame yellow fox | 5 | 15 | 4 |
| 5 the cat likes a fox | 5 | 6 | 6 |
| 6 the fierce cat likes a fox | 6 | 13 | 9 |
| 7 the fierce cat likes a tame fox | 7 | 27 | 19 |
| 8 the little brown cat likes a yellow fox | 8 | 55 | 16 |
| 9 the fierce little brown cat likes a yellow fox | 9 | 111 | 20 |
| 10 the fierce little brown cat likes a tame yellow fox | 10 | 223 | 25 |
| 11 the fierce little brown cat likes a big tame yellow fox | 11 | 447 | 30 |
| 12 the little brown cat likes a big yellow fox |  | 111 | 20 |

Figure 8

## 5. Conclusions

These preliminary results must obviously be interpreted with some caution, since the examples were specially constructed. Further work is in hand to test the performance of the algorithms on larger grammars and more realistic sentences. Because the problem is NP-complete, it is most unlikely that there is an algorithm which will prove efficient in all cases, but the algorithm described here already provides worthwhile improvements in practice, and there is considerable scope for further improvement. For example, for grammars related to HPSG it seems probable that considerable benefit would be gained from adding a constraint propagation component to an unordered version of a head-corner parsing algorithm, as described by Van Noord [Van Noord, 1991]. Alternatively, it may be that constraint graphs, like the LR parsing tables described by Briscoe and Carroll [Briscoe and Carroll, 1991], are suitable locations for the storage of probabilistic information derived from the analysis of corpora

## Acknowledgements

The idea of Shake-and-Bake machine translation is the result of collaboration between Mike Reape and Pete Whitelock. I am grateful to all my colleagues at SLE for providing encouragement and support, but particularly to Pete Whitelock, Andrew Kay, Ian Johnson and Olivier Laurens, who took the trouble to provide detailed commentary. Thanks are also due to two anonymous COLING referees, whose comments produced major revisions.